\documentclass[]{zgca}

\usepackage{amsmath,amssymb}
\usepackage{booktabs}
\usepackage{graphicx}
\usepackage{xspace}
\usepackage{adjustbox}
\usepackage{multirow}
\usepackage{wrapfig}
\usepackage{tabularx}
\usepackage{stfloats}
\usepackage{url}
\usepackage{verbatim}
\usepackage{tikz}
\usepackage{comment}
\usepackage{makecell}
\usepackage{float}

\title{Human-as-Humanoid: Enabling Zero-Shot Humanoid Learning from Ego-Exo Human Videos with Human-Aligned Embodiments}

\author{
Xiaopeng Lin\textsuperscript{1,2,*},
Ruoqi Yang\textsuperscript{2,*},
Shijie Lian\textsuperscript{3,6,*},
Zhaolong Shen\textsuperscript{3,7,*},
Bin Yu\textsuperscript{3,5,*},
Changti Wu\textsuperscript{3},
Haibao Liu\textsuperscript{2},
Yuxiang Zhang\textsuperscript{2},
Hong Li\textsuperscript{2},
Qiyuan Su\textsuperscript{2},
Haochen Liu\textsuperscript{2},
Xuguo He\textsuperscript{2},
Yukun Shi\textsuperscript{4},
Cong Huang\textsuperscript{3,4},
Zhirui Zhang\textsuperscript{2},
Bojun Cheng\textsuperscript{1,$\dagger$},
Kai Chen\textsuperscript{2,3,4,$\dagger$}
\\[0.4em]
\textsuperscript{1}The Hong Kong University of Science and Technology (Guangzhou)
\quad
\textsuperscript{2}DeepCybo
\quad
\textsuperscript{3}ZGCA
\quad
\textsuperscript{4}ZGCI
\\
\textsuperscript{5}Harbin Institute of Technology
\quad
\textsuperscript{6}Huazhong University of Science and Technology
\quad
\textsuperscript{7}Beihang University
\\[0.2em]
\textsuperscript{*}Equal contribution,
\textsuperscript{$\dagger$}Corresponding author
}

\newcommand{\dof}{60-DoF\xspace}
\newcommand{\method}{Human-as-Humanoid\xspace}
\newcommand{\robot}{PrimeU\xspace}

\renewcommand{\paragraph}[1]{\vspace{1.25mm}\noindent\textbf{#1}}

\abstract{
Vision-language-action (VLA) models across robot embodiments require high-quality observation--action supervision to learn deployable action distributions, yet scaling such robot data remains difficult, especially for high-DoF humanoids.
Teleoperation provides controller-aligned supervision, while human egocentric videos capture diverse bimanual manipulation but do not directly provide executable robot actions.
We introduce \method, a human-to-humanoid supervision framework that enables near-real-time human-centric action generation, making human demonstrations usable for high-DoF humanoid VLA training by jointly aligning the robot embodiment, the sensing setup, and the action-label interface.
Built on \robot, a human-aligned \dof upper-body humanoid, \method uses synchronized ego-exo videos to pair deployment-aligned egocentric observations with exocentric motion recovery, retargets the recovered human motion through staged Inverse Kinematics (IK) into controller-aligned \dof action chunks, and trains the VLA model with Forward Kinematics (FK)-aware supervision to preserve wrist and fingertip task-space geometry.
This converts large-scale human demonstrations from visual observations into executable observation--action supervision for the target humanoid.
Experiments validate the conversion chain at the motion-recovery, robot-action-space, and real-robot deployment levels.
\method yields a 4.8--7.2x raw demonstration-throughput gain over humanoid teleoperation in our data-collection analysis, and on several downstream tasks, policies post-trained only with the converted human labels generalize to real-robot deployment without target-task robot demonstrations.
The official project website is available at \url{https://zgc-embodyai.github.io/Human-as-Humanoid/}.
}

\date{\today}
\damodata[Keywords]{Humanoid Manipulation, Human Data, Teleoperation, VLA Models}

\begin{document}
\thispagestyle{firstheader}
\maketitle
\pagestyle{plain}

\let\cite\citep


\section{Introduction}
\label{sec:introduction}

Vision-language-action (VLA) policies provide a unified interface for general robot control by mapping visual observations and language instructions to robot actions~\cite{RT2_2023_CoRL,Octo_2024_arXiv,PI0_2024_arXiv,PI05_2025_arXiv,OpenVLA_2024_CoRL,OpenVLA-OFT_2025_arXiv,GR00T_2025_arXiv}. This formulation makes policy learning depend not only on visual-language representation, but also on the quality and scale of observation-action supervision. Scalable VLA training therefore requires sufficient and reliable observation-action supervision. Such supervision allows the policy to learn an executable action distribution for the target robot.

The granularity and representation of action supervision in VLA learning are coupled to the target robot's embodiment and controller interface. For human-centered manipulation, full-size dexterous humanoids offer a suitable VLA embodiment: their human-scale upper body and dexterous hands support bimanual, contact-rich manipulation~\cite{fu2024humanplus,he2024omnih2o,GR00T_2025_arXiv}. This morphology makes the required labels higher-dimensional and more interface-specific. Wrist or end-effector pose with gripper state remains a compact and effective task-space abstraction, especially for many gripper-based manipulation settings~\cite{chi2024universal}. For dexterous humanoid control, execution also depends on hand preshape, fingertip contact, bimanual coordination, arm-redundancy resolution, and controller-level joint commands~\cite{RDT1B_2025_ICLR,wang2024dexcap,METIS_2025_arXiv}. This setting motivates action labels aligned with the robot execution interface while preserving wrist and fingertip geometry as task-space constraints.

Conventional robot teleoperation is the standard way to obtain such action-interface-aligned supervision: it records observations and actions on the target hardware, with trajectories that already respect the robot controller. The limitation is throughput. Although the robotics community has made major efforts to scale robot data collection~\cite{RT-1_2022_arXiv,Bridgedatav2_2023_CoRL,OXE_2024_ICRA,Droid_2024_arXiv,AgiBotWorld_2025_IROS,fu2024mobile,zhao2024aloha,ding2025bunny}, collecting high-DoF humanoid trajectories directly on hardware remains slow, labor-intensive, constrained by safety, and difficult to diversify across scenes. The core data-efficiency question is therefore how to reduce the dependence on robot-side teleoperation without losing executable action supervision.

Human demonstrations are a natural way to scale this action data~\cite{Ego4D2022_CVPR,EgoExo4D_2024_CVPR,Egocentric100K_2025_Dataset,EgoDex_2025_arXiv,HOI4D_2022_CVPR,Egomimic_2025_ICRA,EgoVLA_2025_arXiv,HRDT_2025_arXiv,Being-H0_2025_arXiv}. Human bimanual manipulation is fast to collect and contains rich behavior, including hand-shape changes, contact timing, regrasping, two-hand coordination, and active visual observation. Egocentric human video is also close to the head-view observation used by humanoid policies. However, raw human video is not robot training data. It contains visual observations and human motion, but it does not contain target-robot action labels; humans and humanoids also differ in body scale, joint structure, hand morphology, degrees of freedom, sensing viewpoints, and reachable workspace. To make human demonstrations useful for VLA action learning, the problem is not simply to collect more videos, but to convert human behavior into robot-learnable and robot-executable supervision.

This conversion exposes four coupled requirements:
\begin{enumerate}
\item[(i)] Embodiment alignment requires the robot morphology and sensing layout to remain compatible with human demonstrations, reducing retargeting error from differences in body scale, reachable workspace, hand morphology, and viewpoints.
\item[(ii)] Observation--motion compatibility requires egocentric streams for deployment-aligned policy inputs and exocentric views for robust upper-body and hand recovery under occlusion.
\item[(iii)] Action-interface alignment requires converted labels to be compatible with the robot's joint ordering, URDF convention, joint limits, and controller interface, in addition to task-space motion intent.
\item[(iv)] Joint--task consistency requires executable joint commands to preserve the wrist and fingertip geometry that is critical for contact-rich manipulation.
\end{enumerate}

To meet these requirements, we introduce \method, a human-to-humanoid supervision pipeline that converts synchronized ego-exo human videos into robot-executable high-DoF labels. The pipeline is built on \robot, a human-aligned \dof upper-body humanoid whose dexterous hands, actuated neck and waist, and head- and wrist-view sensing reduce the embodiment and sensing gap to human bimanual manipulation. During data collection, synchronized ego-exo videos provide both deployment-aligned egocentric observations and exocentric visual evidence for reliable motion recovery. The recovered motion is retargeted through staged IK into controller-aligned \dof action chunks, and FK-aware VLA training preserves wrist and fingertip geometry while learning executable joint actions.

\begin{figure*}[t]
    \centering
    \includegraphics[width=\textwidth]{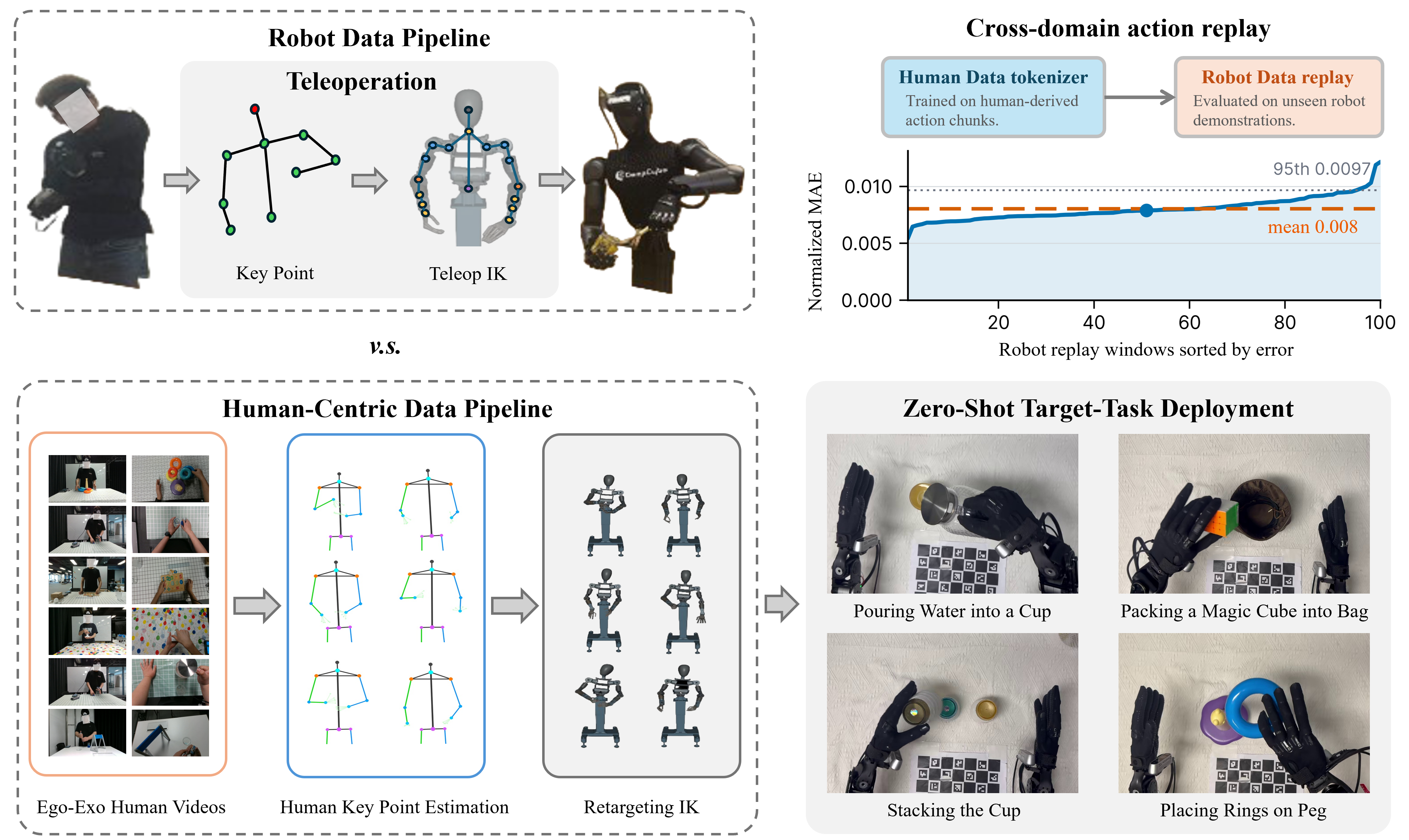}
    \caption{
    Overview of \method and its effect on humanoid action-data generation.
    Conventional robot data collection obtains executable actions through teleoperation, coupling data throughput to robot execution.
    \method instead uses synchronized ego-exo human videos, estimates human keypoints, and retargets them into executable \dof humanoid actions near the video capture rate.
    The upper-right panel reports a cross-domain action-replay compatibility check, where an action tokenizer trained on human-derived action chunks reconstructs unseen robot trajectories.
    The lower-right panel shows zero-shot real-robot policy rollouts from a VLA fine-tuned on converted human demonstrations, not direct replay of generated trajectories, on water pouring, bag packing, cap twisting, and ring placement.
    }
    \label{fig:intro}
\end{figure*}

Figure~\ref{fig:intro} summarizes this shift from robot-side teleoperation to ego-exo human video conversion. The pipeline runs at around 20 FPS and, under a 15 Hz capture setting, yields a 4.8--7.2x raw demonstration-throughput gain over motion-capture teleoperation in our representative data-collection analysis.

We validate the same conversion chain before policy deployment. For observation--motion compatibility, ego-exo camera-only recovery produces more stable human keypoints than motion-capture-suit keypoints in the target manipulation setting. For action-interface alignment, an action tokenizer trained on human-derived robot actions reconstructs unseen robot trajectories with low normalized MAE (0.008 average, 0.0097 at the 95th percentile). For joint--task consistency, a FK-aware VLA policy fine-tuned only on converted human demonstrations deploys on real high-DoF tasks including ring placement, bag packing, cap twisting, and water pouring, without target-task robot demonstrations.

Our contributions are:
\begin{itemize}
    \item We design and use \robot, a human-aligned \dof upper-body humanoid that supports embodiment-aligned conversion from human bimanual manipulation to humanoid action supervision.
    \item We introduce \method, a near-real-time ego-exo video-to-action pipeline that converts camera-only human demonstrations into controller-aligned \dof humanoid action chunks without motion-tracking wearables.
    \item We train high-DoF VLA policies with FK-aware supervision, preserving wrist and fingertip geometry while keeping the policy output in executable joint space.
    \item We validate the conversion chain through keypoint stability, tokenizer reconstruction on unseen robot trajectories, a 4.8--7.2x raw throughput gain, and zero target-task robot-demonstration deployment on real humanoid manipulation tasks.
\end{itemize}


\section{Related Work}
\label{sec:related_work}


\paragraph{Egocentric and ego-exo human data.}
Large-scale human video datasets have made first-person behavior a realistic supervision source for embodied learning. Ego4D~\cite{Ego4D2022_CVPR} provides thousands of hours of daily-life egocentric video and benchmarks for first-person perception, while Ego-Exo4D~\cite{EgoExo4D_2024_CVPR} extends this direction to synchronized first- and third-person views of skilled human activities, and Egocentric-100K~\cite{Egocentric100K_2025_Dataset} further increases the scale of egocentric video resources. More manipulation-oriented resources further move from perception toward action: HOI4D~\cite{HOI4D_2022_CVPR} provides 4D egocentric human-object interaction data, EgoDex~\cite{EgoDex_2025_arXiv} collects large-scale egocentric dexterous manipulation with 3D hand tracking, EgoMimic~\cite{Egomimic_2025_ICRA} uses wearable egocentric video and hand tracking to co-train imitation policies with robot data, ActiveMimic~\cite{ActiveMimic_2026_arXiv} studies egocentric pretraining for active viewpoint control, and EgoScale~\cite{Egoscale_2026_arXiv} shows that dexterous manipulation performance can scale with diverse egocentric human data. These datasets and systems establish the value of human egocentric data. Our work uses the same scaling opportunity, but targets a different output: near-real-time executable \dof upper-body humanoid joint supervision from synchronized camera-only videos, rather than only perception labels, hand trajectories, or pretraining signals.

\paragraph{Robot learning from human videos.}
Recent methods increasingly use human data to reduce the dependence on robot demonstrations. H-RDT~\cite{HRDT_2025_arXiv} pretrains on egocentric human manipulation videos and transfers to robot policies through cross-embodiment action modules and robot fine-tuning. In-N-On~\cite{In-N-On_2025_arXiv} combines in-the-wild and on-task egocentric human data to train language-conditioned manipulation policies. Several recent works further move from semantic video pretraining toward explicit motion or action supervision: EgoVLA~\cite{EgoVLA_2025_arXiv} learns VLA models from egocentric human videos with 3D hand action annotations, Being-H0~\cite{Being-H0_2025_arXiv} scales VLA pretraining with human hand-motion representations, VITRA~\cite{VITRA_2025_arXiv} converts real-life human activity videos into robot-aligned action segments, and MotionTrans~\cite{MotionTrans_2025_CoRLWorkshop} studies motion-level learning from human VR demonstrations. Ego-Pi~\cite{EgoPi_2026_CVPR} uses ego-centric human and robot data jointly, showing that human videos can contribute high-level task semantics such as ordering, sorting rules, and skill composition for dexterous humanoids; its demonstrated transfer is therefore complementary to ours, which learns target-task low-level action labels without target-task robot demonstrations. PhysBrain~\cite{PhysBrain_2025_arXiv} and PhysBrain 1.0~\cite{PhysBrain_1.0_2026_arXiv} translate human egocentric videos into physical commonsense, planning, state-tracking, and interaction supervision for VLM/VLA adaptation. EgoEngine~\cite{EgoEngine_2026_arXiv} converts Aria egocentric human videos into paired robot observations and executable dexterous trajectories through digital-twin reconstruction, robot-video synthesis, and object-centric simulation optimization. HumanEgo~\cite{HumanEgo_2026_arXiv} takes a strong robot-data-free direction by learning from minutes of human egocentric video through hand-object interaction representations, but its action interface is a virtual gripper rather than a dexterous humanoid upper body. In contrast, \method focuses on the action-label generation problem for humanoids: it converts synchronized camera-only human videos into robot-controller-aligned actions over arms, dexterous hands, neck, and waist, runs at around the video capture rate, and supports real-robot deployment without target-task robot demonstrations.




\section{Human-Aligned Humanoid Embodiment}
\label{sec:robot_embodiment}

Our system starts from the robot embodiment rather than treating human-to-robot transfer as a purely post-hoc retargeting problem. If the target robot differs strongly from the human in reach, sensing viewpoint, or dexterous action structure, then human demonstrations must cross a large embodiment gap before they can supervise a policy. \robot is designed to reduce this gap at the source. Its upper body follows standard adult-male manipulation proportions, including overall height, shoulder width, and arm length, so that common human demonstrations occupy a workspace close to the robot's reachable workspace.

\begin{figure}[t]
    \centering
    \includegraphics[width=\linewidth]{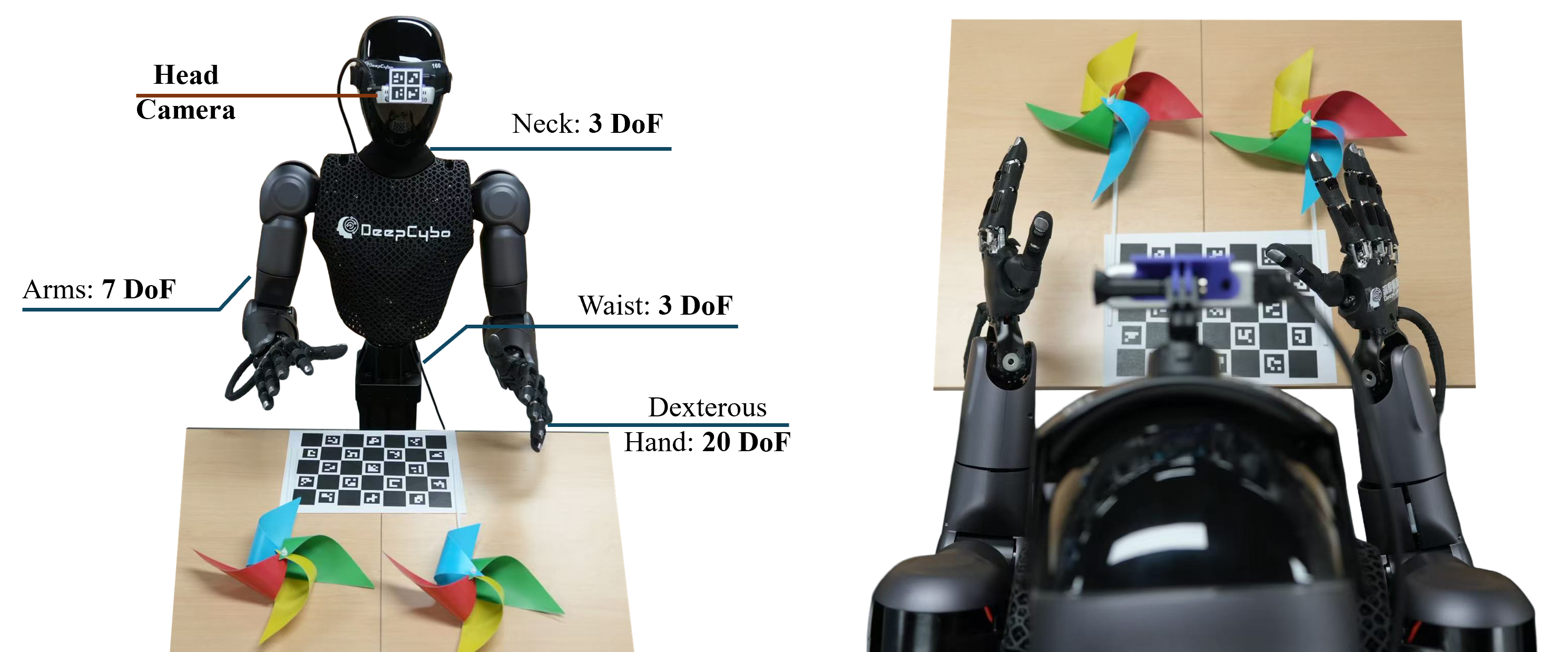}
    \caption{
    \robot human-aligned humanoid embodiment.
    The upper body is developed in-house and is designed around adult-male manipulation scale, with two 7-DoF arms, a 3-DoF neck, and a 3-DoF waist.
    The platform uses two 20-DoF Wuji dexterous hands for fine manipulation.
    Head-view and wrist-view cameras are Intel RealSense D435 cameras, matching the viewpoint structure used by the deployed VLA policy.
    }
    \label{fig:robot_setup}
\end{figure}

Table~\ref{tab:primeu_anthro} makes this human alignment numerically checkable. We compare action-relevant upper-body dimensions of \robot against adult-male anthropometric references from ANSUR II~\cite{Gordon2014ANSURII}. The goal is not to exactly reproduce human anatomy, but to keep the robot's shoulder span, reach length, and hand scale close to a common human manipulation scale. The shoulder-to-head measurement is included to characterize the head-camera mounting height; it is less central to manipulation reach than shoulder breadth, arm reach, and hand length, which are all close to human scale.

\paragraph{Kinematic alignment.}
\robot exposes a unified \dof upper-body action space over arms, dexterous hands, neck, and waist. The command interface contains 14 arm joints, 40 hand joints, 3 neck joints, and 3 commanded waist DoFs. Each arm follows a serial shoulder-pitch/roll/yaw, elbow-pitch, wrist-roll/pitch/yaw chain, providing human-like reaching redundancy. Each Wuji hand is modeled as five four-joint finger chains mounted to a palm, with explicit fingertip links that are later used as FK endpoints for geometric supervision. The actuated neck controls the sensor-bearing head, and the actuated waist expands the reachable workspace for torso-assisted manipulation. This design makes the robot action space closer to the structure of human upper-body motion than a low-DoF gripper interface, while still remaining a directly executable joint-space controller interface.

\paragraph{Sensing alignment.}
The robot uses head-view and wrist-view RealSense D435 cameras. The head-view camera provides an egocentric view of the workspace, while wrist-view cameras capture hand-centric contact and object geometry. These viewpoints mirror the human camera streams used for policy inputs, so the visual observations seen during deployment are aligned with the observation structure used during human-data training.

\paragraph{Action-interface alignment.}
Human alignment alone does not make human data executable. We expose the robot to learning as one joint-space action vector over the controllable upper-body groups, rather than mixing human pose, end-effector pose, and virtual-gripper commands. Human videos are therefore converted into \robot action chunks rather than generic human-motion annotations. This action-level interface is the foundation that lets \method treat human demonstration as a scalable source of robot-action supervision.

\begin{table}[t]
    \centering
    \small
    \caption{
    Action-relevant anthropometric scale alignment.
    Human values are 50th-percentile male measurements from ANSUR II~\cite{Gordon2014ANSURII}; \robot values are computed from the URDF kinematic tree and visual meshes.
    Human shoulder-to-head height is sitting height minus sitting acromial height.
    For \robot, shoulder breadth is the shoulder-roll joint-center span, shoulder-to-head height is measured from the shoulder-roll level to the head visual top, and reach is the shoulder-roll-to-palm kinematic chain plus the palm-to-middle-fingertip link.
    }
    \label{tab:primeu_anthro}
    \begin{tabularx}{\linewidth}{@{}Xccc@{}}
        \toprule
        Dimension & Human (cm) & \robot (cm) & Ratio \\
        \midrule
        Shoulder breadth & 41.5 & 40.4 & 0.97 \\
        Shoulder-to-head height & 31.5 & 37.1 & 1.18 \\
        Shoulder-to-middle-fingertip reach & 78.6 & 80.3 & 1.02 \\
        Hand length & 19.3 & 19.3 & 1.00 \\
        \bottomrule
    \end{tabularx}
\end{table}


\section{Near-Real-Time Human-Centric Action Generation}
\label{sec:data_construction}

Given \robot as the target embodiment, we convert synchronized camera-only human demonstrations into executable humanoid action labels at collection time. A head-mounted egocentric camera provides the policy observation. Optional wrist cameras can be recorded as extra policy inputs, but they are not used for pose tracking. One or more exocentric RGB views provide stable evidence for human body and hand recovery. Since all streams are time-aligned, each egocentric observation can be paired with the corresponding upper-body motion. The conversion runs at about 20 FPS in our implementation, which is close to a common 15 Hz capture rate.

The label-construction pipeline has four stages. First, we track the demonstrated person in the exocentric video and propagate the target mask over short temporal windows. Second, a mesh-aware human reconstruction module recovers the upper-body pose and hand motion from the tracked video. We use the reconstructed mesh only as a motion-estimation intermediate. The stored representation is a compact camera-coordinate skeleton with upper-body and hand keypoints. Third, the skeleton is smoothed in a root-relative coordinate frame and mapped into the selected kinematic convention. This step aligns axes, interpolates torso and neck joints, and builds palm frames for both hands. Finally, a staged IK solver retargets the skeleton to \robot and emits controller-aligned \dof labels.

The output is not a human-pose annotation. It is a robot training tuple:
\[
    (o_t,\ell,q_t,q^{\ast}_{t+1:t+H}),
\]
where $o_t$ is the egocentric observation, $\ell$ is the language instruction, $q_t$ is the current robot state, and $q^{\ast}_{t+1:t+H}$ is a future robot action chunk, both $q_t$ and $q^{\ast}_{t+1:t+H}$ are robot states obtained by retargeting the recovered human motion to PrimeU joint space. The goal is to turn human motion into the same joint space used by policy training and robot control.

The robot label is a unified upper-body humanoid joint vector
\begin{equation}
    q =
    \left[
        q^{L}_{arm},
        q^{L}_{hand},
        q_{neck},
        q_{waist},
        q^{R}_{arm},
        q^{R}_{hand}
    \right] \in \mathbb{R}^{60},
\end{equation}
which contains two 7-DoF arms, two 20-DoF Wuji dexterous hands, a 3-DoF neck, and a 3-DoF waist. This single joint-space convention is used for retargeting, learning, and deployment. The hands preserve fine finger articulation. The neck controls the egocentric sensor pose. The waist expands the reachable workspace and supports torso-assisted reaching.

Retargeting is solved in stages because different robot parts require different target geometry. A monolithic \dof IK problem couples fingers, wrists, arms, neck, and waist, and it is poorly conditioned. We instead define a target vector $y_b^{\ast}$ and a robot kinematic map $f_b$ for each body part $b$, then solve a regularized IK problem
\begin{equation}
    q_b^{\star}
    =
    \arg\min_{q_b\in\mathcal{Q}_b}
    \left\|
        W_b\left(f_b(q_b;\bar{q})-y_b^{\ast}\right)
    \right\|_2^2
    +
    \lambda_b\left\|q_b-q_b^{0}\right\|_2^2 ,
    \label{eq:part_ik}
\end{equation}
where $\mathcal{Q}_b$ denotes joint limits, $\bar{q}$ contains already-fixed upstream joints, and $q_b^0$ is the seed or previous solution. Hands are solved first. The targets include calibrated fingertip points and finger-segment directions. For an observed human fingertip $x_i^h$, the target is aligned by a calibrated similarity transform $\mathcal{T}_{h\rightarrow r}$, producing residuals of the form
\begin{equation}
    e_i^{hand}(q_b)
    =
    p_i^{r}(q_b)
    -
    \mathcal{T}_{h\rightarrow r}(x_i^h),
\end{equation}
which are minimized by a seed retargeter followed by per-finger Levenberg--Marquardt refinement. Arm motion is then solved with wrist, palm, and elbow targets using damped Jacobian IK. Wrist orientation is extracted from the hand frame and refined after the arm solve, so the final palm pose remains consistent with the shoulder-elbow chain. Neck and waist motion use semantic head and torso frames rather than point targets. We minimize an $SO(3)$ residual
\begin{equation}
    e^{ori}(q_b)
    =
    \mathrm{Log}\!\left(
        R_b(q_b)^{\top}R_b^{\ast}
    \right),
\end{equation}
with temporal regularization. The order is therefore hand, arm and wrist, neck and waist, then guard and smooth.

After the per-part solves, the trajectory is filtered by joint limits, velocity bounds, and acceleration bounds. Local wrist-palm refinements are accepted only when they improve the relevant task-space residual over the current baseline:
\begin{equation}
    q^{new}
    =
    \begin{cases}
    \tilde{q}, & E_b(\tilde{q}) < E_b(q^{base})-\epsilon,\\
    q^{base}, & \text{otherwise},
    \end{cases}
    \label{eq:guarded_accept}
\end{equation}
where $E_b$ is the residual used by the corresponding body-part solver. This guarded rule prevents local solver failures from degrading otherwise good labels.

The result is a robot-centric action corpus rather than a human-motion corpus. The same URDF and joint ordering are used by the IK retargeting pipeline, the differentiable FK layer during training, and the robot controller at deployment. Wrist positions, fingertip positions, and joint limits therefore have the same meaning throughout the system.


\section{PhysDex: VLA Learning from Human-Derived High-DoF Actions}
\label{sec:method}

\subsection{Formulation}

After human-to-humanoid retargeting, each segment is represented as $(o_t,\ell,q_t,q^{\ast}_{t+1:t+H})$. Here $o_t$ is the egocentric image, $\ell$ is the language instruction, $q_t\in\mathbb{R}^{60}$ is the current robot state, and $q^{\ast}_{t+1:t+H}$ is the future robot joint trajectory. The policy predicts future-relative joint offsets. For FK and pose losses, these offsets are decoded back to absolute joints:
\begin{equation}
    \hat{q}_{t+h}=q_t+\hat{A}_h,
    \qquad
    q^{\ast}_{t+h}=q_t+A^{\ast}_h,
    \quad h=1,\ldots,H .
\end{equation}
In our implementation, the chunk contains 40 future states and each state follows the same \dof ordering used by the IK labels.

The action is represented in joint space rather than as end-effector poses. This choice is central to the method. It avoids deployment-time IK, preserves the null-space structure of the multi-finger hands, and makes neck and waist motion part of the same action convention as the arms and hands. The cost of this choice is that plain joint-space supervision is poorly aligned with dexterous task geometry. The remainder of the method addresses this cost without changing the executable output space.

\begin{figure*}[t]
    \centering
    \includegraphics[width=\textwidth]{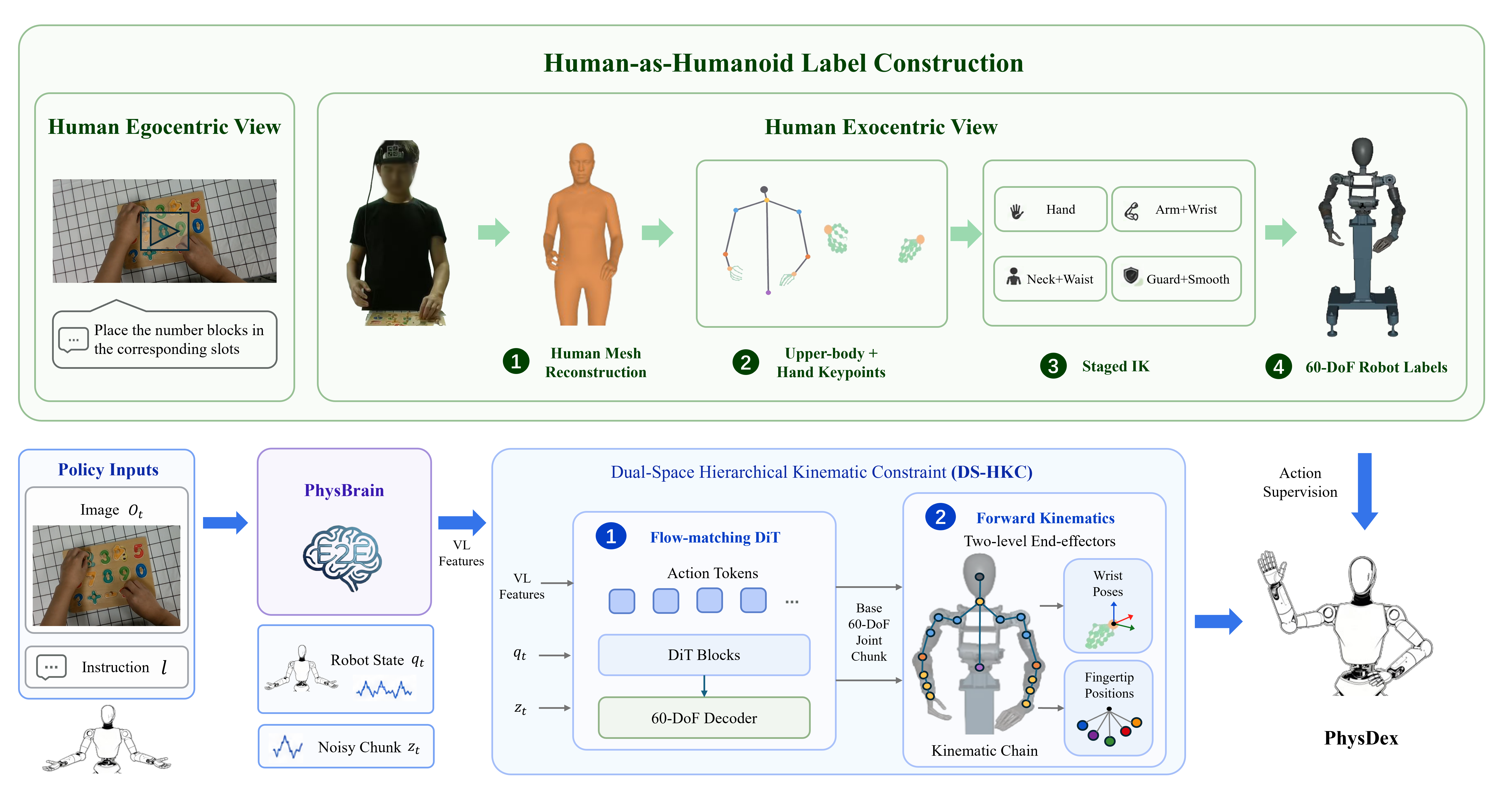}
    \caption{Overview of the PhysDex learning pipeline. (Top) Synchronized human ego-exo videos are converted into controller-aligned \dof robot action labels: egocentric views provide deployment-aligned policy observations, exocentric views support human mesh and upper-body/hand keypoint recovery, and staged IK maps the recovered motion into \robot joint chunks. (Bottom) The VLA policy is trained in the same robot action space. Visual-language features, the current robot state, and a noisy action chunk condition a flow-matching DiT, while DS-HKC applies differentiable FK supervision to the predicted joint chunk through wrist poses and fingertip positions.}
    \label{fig:method_overview}
\end{figure*}

The policy-learning stack receives the image observation, language instruction, current robot state, and a noisy action chunk. A VLM encodes the visual-language context. A flow-matching DiT predicts future \dof joint actions. Dual-Space Hierarchical Kinematic Constraint (DS-HKC) then supervises the same prediction in task space through differentiable FK. Figure~\ref{fig:method_overview} summarizes this two-stage conversion and learning pipeline.

\subsection{Action Backbone}

The egocentric image and instruction are encoded by PhysBrain, which we use as an existing manipulation-aware VLM backbone. PhysBrain supplies visual-language tokens for objects, hands, contact cues, and task progress. Proprioception and diffusion noise enter the action model directly. They are not routed through the VLM, which keeps the perceptual prior separate from the action-generation mechanism.

The action backbone is a conditional flow-matching DiT. Let $h_{\phi}=E_{\phi}(o_t,\ell)$ denote the visual-language tokens, $A^{\ast}$ the future-relative target chunk, and $z_0\sim\mathcal{N}(0,I)$ a noise chunk. For an interpolation time $\tau$, the model is trained on the linear path $z_{\tau}=(1-\tau)z_0+\tau A^{\ast}$ with target velocity $A^{\ast}-z_0$:
\begin{equation}
    \mathcal{L}_{fm}
    =
    \mathbb{E}_{A^{\ast},z_0,\tau}
    \left[
    \left\|
    v_{\theta}(z_{\tau},\tau,q_t,h_{\phi})
    -
    (A^{\ast}-z_0)
    \right\|_2^2
    \right].
    \label{eq:flow_matching_loss}
\end{equation}
At inference, integrating the velocity field yields a future-relative \dof action chunk. The chunk is decoded through the robot kinematic convention and sent directly to the controller. We use four denoising steps at inference.

\subsection{Dual-Space Hierarchical Kinematic Constraint}

High-DoF humanoid action learning has a space mismatch. The policy must output executable joint commands, but manipulation success is often decided by wrist motion, fingertip placement, and contact geometry. DS-HKC addresses this mismatch with supervision in two spaces. Joint-space losses keep the output close to the demonstrated \dof trajectory. FK-space losses supervise the terminal geometry induced by that trajectory.

The FK map is induced by the \robot URDF and uses the same joint order as retargeting and deployment. Given an absolute joint state $q$, it returns wrist positions $W(q)$, wrist rotations $R(q)$, and fingertip positions $P(q)$:
\begin{equation}
    \mathcal{F}_{\mathcal{U}}(q)=
    \left(
        W(q), R(q), P(q)
    \right).
\end{equation}
Here $W(q)\in\mathbb{R}^{2\times3}$ covers the two wrists, $R(q)\in\mathbb{R}^{2\times3\times3}$ covers their orientations, and $P(q)\in\mathbb{R}^{2\times5\times3}$ covers five fingertips on each hand.

DS-HKC supervises two end-effector levels as a hierarchical architecture. The wrist level constrains the proximal hand pose. The fingertip level constrains distal contact geometry. For a predicted chunk $\hat{A}$ and target chunk $A^{\ast}$, we decode both to absolute trajectories and compute
\begin{equation}
\begin{aligned}
    \mathcal{L}_{wrist}
    &=
    \frac{1}{H}
    \sum_{h=1}^{H}
    \left[
    \left\|
        W(\hat{q}_{t+h})-W(q^{\ast}_{t+h})
    \right\|_2^2
    +
    \lambda_R
    \left\|
        R(\hat{q}_{t+h})-R(q^{\ast}_{t+h})
    \right\|_F^2
    \right],
    \\
    \mathcal{L}_{tip}
    &=
    \frac{1}{H}
    \sum_{h=1}^{H}
    \left\|
        P(\hat{q}_{t+h})-P(q^{\ast}_{t+h})
    \right\|_2^2 .
\end{aligned}
\end{equation}
The DS-HKC term aggregates wrist-level geometry, fingertip-level geometry, and joint-limit feasibility:
\begin{equation}
    \mathcal{L}_{dshkc}
    =
    \lambda_{wrist}\mathcal{L}_{wrist}
    +
    \lambda_{tip}\mathcal{L}_{tip}
    +
    \lambda_{lim}\mathcal{L}_{lim}.
    \label{eq:fk_loss}
\end{equation}
These task-space quantities do not require additional annotations; they are computed by applying the same URDF-induced FK map to predicted and target joint trajectories. Let $f_{\mathcal{U}}$ denote the FK map from robot joints to the selected wrist and fingertip sites. For a local squared task-space error, the gradient with respect to the joint configuration is
\begin{equation}
    \nabla_q \|f_{\mathcal{U}}(q)-f_{\mathcal{U}}(q^\ast)\|_2^2
    =
    2J(q)^{\top}\left(f_{\mathcal{U}}(q)-f_{\mathcal{U}}(q^\ast)\right),
\end{equation}
where $J(q)$ is the manipulator Jacobian. Under a local linearization, this objective weights joint-space errors through the configuration-dependent matrix $J(q)^{\top}J(q)$. Joint deviations that produce larger wrist or fingertip displacement receive stronger corrective gradients, making DS-HKC complementary to joint-space regression for contact-rich manipulation.

\subsection{Training Objective}

The objective combines flow matching, decoded action losses, trajectory smoothness, and DS-HKC. The flow-matching term $\mathcal{L}_{fm}$ is defined in Eq.~\eqref{eq:flow_matching_loss}. The decoded action terms supervise absolute future pose, step-wise delta, and temporal smoothness. DS-HKC supervises wrist geometry, fingertip geometry, and joint limits through the robot kinematic model.
The full objective is
\begin{equation}
\begin{aligned}
    \mathcal{L}
    =
    &\lambda_{fm}\mathcal{L}_{fm}
    +
    \lambda_{abs}\mathcal{L}_{abs}
    +
    \lambda_{\Delta}\mathcal{L}_{\Delta}
    +
    \lambda_{sm}\mathcal{L}_{sm}
    +
    \alpha(s)\mathcal{L}_{dshkc},
\end{aligned}
\label{eq:total_objective}
\end{equation}
where $\lambda_{fm}$, $\lambda_{abs}$, $\lambda_{\Delta}$, and $\lambda_{sm}$ weight the flow-matching, absolute-pose, action-delta, and smoothness terms, respectively. $\alpha(s)$ denotes the warm-up coefficient for the DS-HKC term at training step $s$.


\section{Experiments}
\label{sec:experiments}

The experiments evaluate whether the proposed ego-exo Human-as-Humanoid data-processing pipeline provides reliable supervision for PhysDex, our high-DoF humanoid VLA policy. We follow the four requirements introduced in Section~\ref{sec:introduction}: embodiment alignment, observation--motion compatibility, controller-aligned action labels, and task-space geometry under joint-space execution. Concretely, we ask:
\begin{enumerate}
    \item Does the target embodiment reduce the human-to-robot morphology gap enough for stable conversion into executable robot actions?
    \item Is camera-only ego-exo motion recovery stable enough for upper-body and hand supervision when compared with a wearable motion-capture suit?
    \item Do human-derived \dof action chunks remain compatible with unseen real \robot trajectories?
    \item Does FK-aware training in PhysDex improve task-space wrist and fingertip geometry while preserving executable high-DoF joint actions and supporting target-task deployment?
\end{enumerate}

\subsection{Experimental Setup and Data Protocol}

Each policy receives visual observations, a language instruction, and the current \dof robot state, and predicts a future action chunk. Human-derived data is collected with synchronized egocentric and exocentric cameras, converted into upper-body and hand keypoints, and retargeted through staged IK into \robot action labels. The egocentric stream matches the policy-observation perspective, while the exocentric stream supplies less occluded evidence for motion recovery.

Embodiment alignment is evaluated as a prerequisite for the conversion chain. Section~\ref{sec:robot_embodiment} and Table~\ref{tab:primeu_anthro} show that \robot is close to human manipulation scale in shoulder breadth, arm reach, and hand length, while also characterizing the shoulder-to-head geometry that determines the head-view sensing height. This design reduces the extrapolation required by retargeting: recovered human wrist, hand, neck, and torso motion can be mapped into the robot's reachable workspace and \dof action structure by staged IK, rather than being interpreted only as generic human pose. The end-to-end replay in Figure~\ref{fig:ego_exo_compare} then checks the action-level consequence of this alignment by verifying that the converted commands preserve the main phases of bimanual manipulation in simulation and on the real robot.

The wrist-view assumption is stage-specific. The broad human-video pretraining stage does not require wrist-camera input, since wrist views are not available for all human sequences. In the post-training and robot-aligned evaluation stage, the observation streams are collected under the same head- and wrist-view convention as the robot data, so the policy interface remains consistent with deployment.

Robot experiments use the \robot humanoid described in Section~\ref{sec:robot_embodiment}. Downstream tasks cover object placement, container manipulation, pouring, cup stacking, tool use, and rotational dexterous manipulation: measuring object temperature with a thermometer gun, placing rings on pegs, loosening and removing a light bulb, packing a magic cube into a bag, pouring water from a cold-water pitcher into a cup, loosening and removing a bottle cap, and stacking cups. We report the stage completion rates and final task success rate. Lower is better for action and geometry metrics. For real-robot deployment, stage completion measures whether each ordered subgoal in Table~\ref{tab:task_stages} is reached within a rollout, while final task success requires completing the full task objective.

\begin{table*}[t]
\centering
\small
\caption{Stage-wise evaluation protocol for real-robot manipulation tasks. Stage completion rates are computed for the ordered subgoals listed below; final task success is counted only when the full task objective is achieved.}
\label{tab:task_stages}
\begingroup
\newcommand{\evalstage}[2]{\hspace*{1.0em}\textbf{S#1.}~#2\par}
\renewcommand{\arraystretch}{1.15}
\hspace*{0.035\textwidth}%
\begin{tabularx}{0.93\textwidth}{@{}>{\raggedright\arraybackslash}p{0.19\textwidth}@{\hspace{1.9em}}>{\raggedright\arraybackslash}X@{}}
\toprule
Task & Ordered evaluation stages \\
\midrule
\textbf{Temperature-gun measurement} &
\evalstage{1}{Right hand approaches and aligns with the thermometer gun.}
\evalstage{2}{Right hand grasps the thermometer gun.}
\evalstage{3}{Left hand approaches and aligns with the object.}
\evalstage{4}{Left hand grasps the object.}
\evalstage{5}{Right hand aims at the object and presses the button.} \\[0.3em]
\textbf{Ring placement} &
\evalstage{1}{Hand approaches the ring.}
\evalstage{2}{Hand correctly lifts the ring.}
\evalstage{3}{The ring is successfully placed onto the peg.} \\[0.3em]
\textbf{Light-bulb loosening} &
\evalstage{1}{Left hand stabilizes the bulb base.}
\evalstage{2}{Right hand approaches and grasps the bulb.}
\evalstage{3}{Right hand performs a twisting motion.}
\evalstage{4}{The bulb is loosened and removed.} \\[0.3em]
\textbf{Magic-cube packing} &
\evalstage{1}{Right hand approaches the cube.}
\evalstage{2}{Right hand grasps the cube.}
\evalstage{3}{The cube is successfully placed into the bag.} \\[0.3em]
\textbf{Water pouring} &
\evalstage{1}{Hand aligns with the kettle handle.}
\evalstage{2}{Hand successfully grasps the handle.}
\evalstage{3}{The robot lifts the kettle and initiates a pouring motion.}
\evalstage{4}{Water is poured into the cup.} \\[0.3em]
\textbf{Bottle-cap loosening} &
\evalstage{1}{Left hand approaches and grasps the bottle.}
\evalstage{2}{Right hand approaches and aligns with the cap.}
\evalstage{3}{Right hand performs a twisting motion.}
\evalstage{4}{The cap is loosened and removed.} \\[0.3em]
\textbf{Cup stacking} &
\evalstage{1}{Left hand aligns with and grasps the left cup.}
\evalstage{2}{The left cup is stacked onto the center cup.}
\evalstage{3}{Right hand aligns with and grasps the right cup.}
\evalstage{4}{The right cup is stacked onto the center cup.} \\
\bottomrule
\end{tabularx}
\endgroup
\end{table*}

\begin{figure*}[t]
    \centering
    \includegraphics[width=0.95\textwidth]{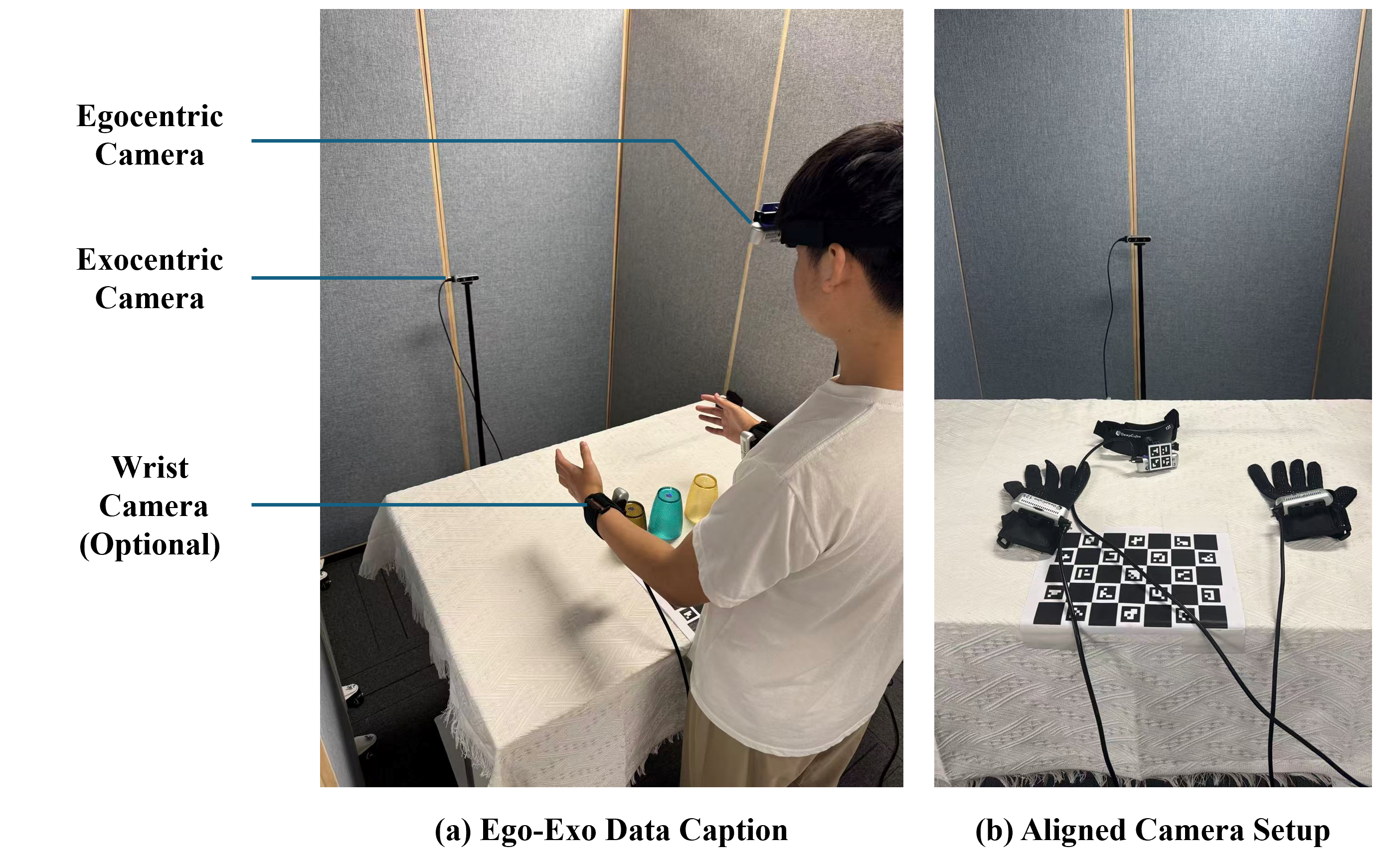}
    \caption{Ego-exo data-collection setup. The egocentric cameras provide policy-aligned observations, while synchronized exocentric cameras provide complementary visual evidence for upper-body and hand motion recovery.}
    \label{fig:ego_exo_capture_setup}
\end{figure*}

\subsection{Ego-Exo Motion Recovery}

This validation corresponds to the observation--motion compatibility requirement in Section~\ref{sec:introduction}. The question is whether camera-only ego-exo data can provide reliable motion recovery while preserving the egocentric observation used by the policy. Figure~\ref{fig:ego_exo_capture_setup} shows the synchronized capture setup used to separate these two roles: egocentric cameras provide policy-aligned observations, while exocentric cameras provide less occluded evidence for motion recovery. We evaluate this requirement with a projection-based diagnostic. For each sequence, we project the skeletons recovered by the ego-exo Human-as-Humanoid pipeline and by the wearable motion-capture system back onto the human video. Image-plane alignment with the visible body provides an intuitive check of recovery accuracy for the upper body, wrists, and hands, which are the components later mapped to \robot joints.

\begin{figure*}[t]
    \centering
    \includegraphics[width=0.95\textwidth]{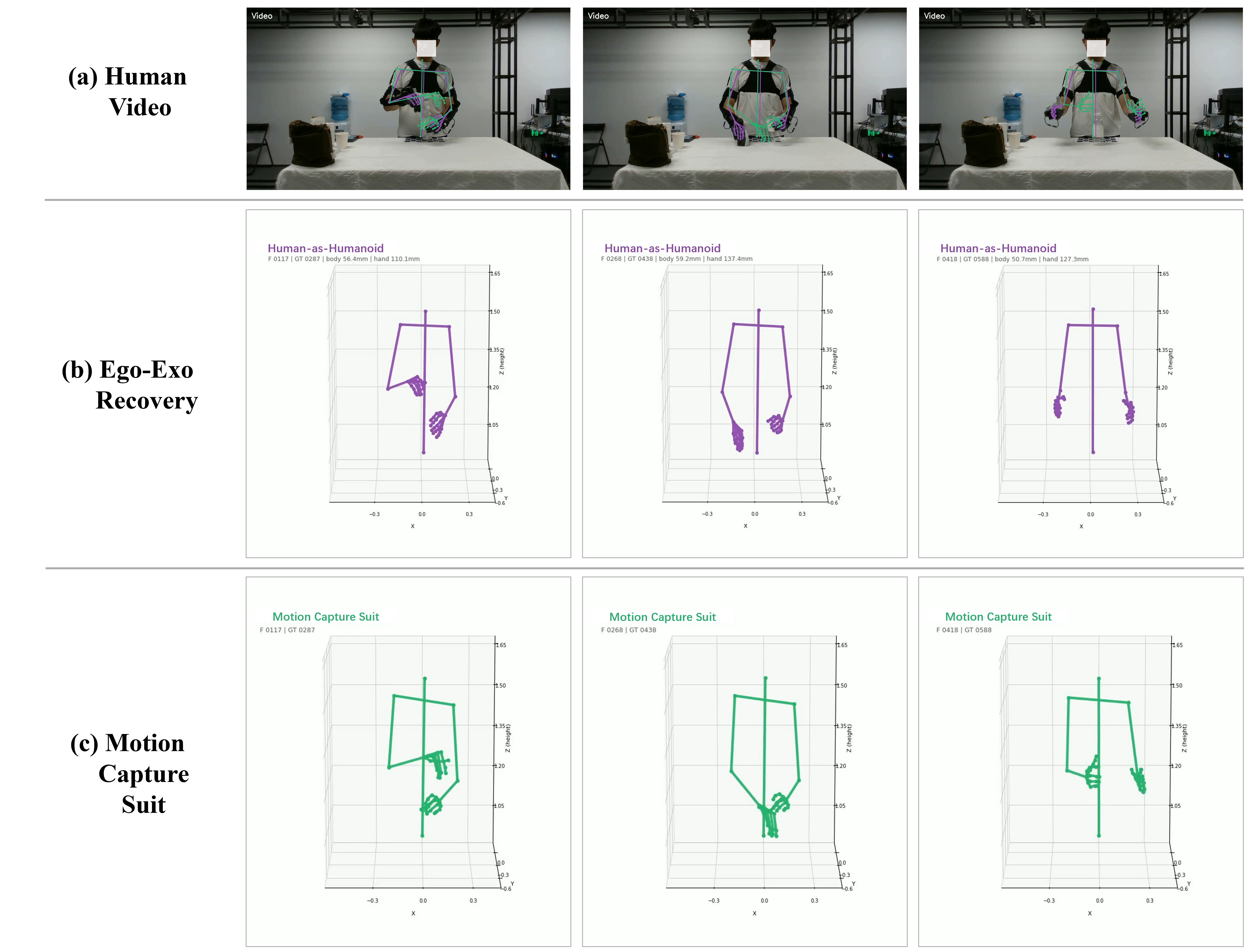}
    \caption{Projection-based comparison between human video, camera-only ego-exo motion recovery, and wearable motion capture at matched relative timestamps. The visualization evaluates observation--motion compatibility by showing how the recovered skeletons align with the visible human body before retargeting into robot action labels.}
    \label{fig:motion_recovery_comparison}
\end{figure*}

Figure~\ref{fig:motion_recovery_comparison} shows that the wearable inertial capture result exhibits visible localization drift in the projected view. This behavior is consistent with the known sensitivity of inertial capture to accumulated global-position error and calibration quality. The effect is especially relevant for close-range bimanual manipulation, where wrist and hand positions must remain aligned with objects. In data collection, this drift often requires operator compensation or repeated recalibration. Both interventions restrict the range of natural demonstrations and reduce effective collection throughput.

Under the same actions, the ego-exo recovery maintains closer projected alignment with the observed body and hands. The exocentric views provide stable geometric evidence for motion recovery, while the egocentric stream remains available as the deployment-aligned policy input. This directly addresses the observation--motion conflict: the system can recover manipulation-relevant joints without replacing the policy observation with an exocentric view. The recovered motion is then retargeted through the \robot URDF, joint limits, and staged IK solver, so this recovery stage supports low-cost and high-precision generation of robot action labels.

\begin{figure*}[t]
    \centering
    \includegraphics[width=0.98\textwidth]{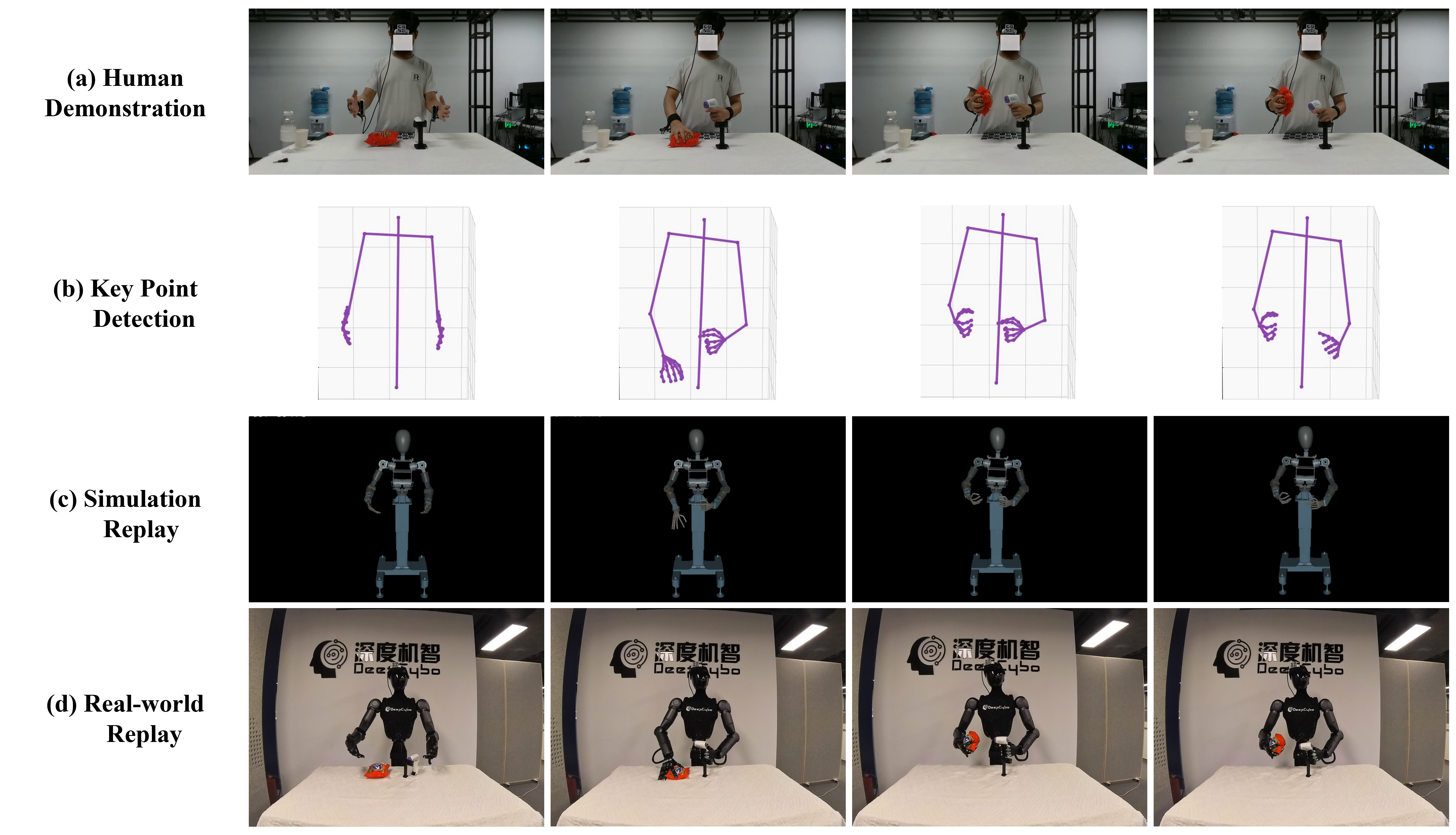}
    \caption{End-to-end ego-exo conversion and replay at matched timestamps. Rows show the source human video, recovered skeleton keypoints, simulator replay, and real-robot replay. The visualization evaluates whether the recovered human motion remains coherent after retargeting into \robot actions and whether the resulting commands preserve the temporal structure required for near-real-time teleoperation-style replay.}
    \label{fig:ego_exo_compare}
\end{figure*}

We further evaluate the full conversion chain after motion recovery. Figure~\ref{fig:ego_exo_compare} traces matched timestamps from the source human video to recovered skeletal keypoints, simulator replay, and real-robot replay. The four-row comparison tests whether the recovered motion remains consistent after retargeting into \robot joint commands, rather than only after projection onto the human image. The simulator and real-robot rows preserve the main bimanual motion phases, including approach, grasp, object transfer, and release. This qualitative consistency supports the accuracy of the data-processing pipeline used to generate high-DoF action labels. Since the same pipeline runs near the video capture rate, it also supports near-real-time teleoperation-style replay for data collection and system debugging.

\subsection{Action-Interface Compatibility}

This experiment addresses the third question in our evaluation: whether human-derived \dof action chunks are compatible with the action space used by real \robot trajectories. Motion recovery and simulator replay verify that human motion can be recovered and retargeted, but motion-level consistency alone is insufficient to establish compatibility with the robot action interface used for policy learning. As illustrated by the upper-right panel of Figure~\ref{fig:intro}, we therefore use a cross-domain action-replay test. A discrete action tokenizer is trained on action chunks generated from human data, using the same \robot joint ordering, kinematic convention, joint limits, and controller interface as deployment. We then evaluate whether this tokenizer can encode and decode held-out real-robot trajectories that were not used to train the tokenizer. If the converted labels formed an interface-specific distribution far from real robot actions, the tokenizer would reconstruct these held-out robot trajectories with large distortion.

We evaluate action-interface compatibility with two complementary tokenizer diagnostics, summarized in Table~\ref{tab:action_interface_compatibility}. First, we train a single \dof tokenizer only on human-derived robot-action chunks and evaluate it on held-out real-robot action windows. This is a strict cross-domain test: the tokenizer never observes robot trajectories during fitting, but must still discretize and reconstruct them under the same \robot action convention. On 100 real-robot windows, the human-only tokenizer reaches a mean normalized MAE of 0.0080 and a 95th-percentile normalized MAE of 0.0097. This low cross-domain reconstruction error indicates that the actions produced by our human-to-\robot conversion occupy a normalized action manifold close to real \robot demonstrations.

We further measure the effect of action reconstruction in task space by applying the original and decoded actions to the \robot kinematic model and comparing the resulting left- and right-hand end-effector trajectories. The human-only tokenizer reconstructs unseen robot windows with a mean bi-manual end-effector error of 5.34 mm, showing that the cross-domain action error remains small after forward kinematics. We use robot-only and robot+human-data tokenizers with the same single-\dof architecture as complementary replay diagnostics for the effect of adding human-derived actions. A tokenizer trained only on robot data reaches a mean bi-manual end-effector error of 4.09 mm, while the tokenizer trained on robot data plus full human-derived actions reaches 4.86 mm. Normalized MAE is reported to show that the cross-domain action-space discrepancy is small, but it is not used to rank the cross-domain and in-domain settings because it also depends on the action variation scale and the normalization induced by each training distribution. We therefore use end-effector error as the primary replay-quality metric. Under this task-space metric, the robot-only tokenizer gives the lowest error, while the human-only and mixed-domain tokenizers still maintain millimeter-scale replay accuracy. Together, these diagnostics support that our staged-IK conversion produces human-derived labels that are compatible with the real \robot action interface.

\begin{table}[t]
    \centering
    \small
    \caption{
    Action-interface compatibility diagnostics.
    All settings use a single \dof tokenizer and 100 real-robot evaluation windows.
    The human-only tokenizer is evaluated cross-domain, while the robot-only and robot+human-data tokenizers provide complementary replay diagnostics for the effect of adding human-derived actions.
    End-effector error is computed by forward kinematics on the decoded action sequence and averaged over the left and right hand sites.
    Normalized MAE is reported as an action-space compatibility diagnostic, not as a strict ranking criterion across settings.
    Lower is better for both metrics.
    }
    \label{tab:action_interface_compatibility}
    \setlength{\tabcolsep}{4.5pt}
    \renewcommand{\arraystretch}{1.08}
    \begin{tabular*}{\linewidth}{@{\extracolsep{\fill}}llccc@{}}
        \toprule
        \multirow{2}{*}{Diagnostic} & \multirow{2}{*}{Training data} & \multirow{2}{*}{Eval.} & EE error (mm) & Norm. MAE \\
        & & & mean / p95 & mean / p95 \\
        \midrule
        Cross-domain & Human only & Robot & 5.34 / 12.67 & 0.0080 / 0.0097 \\
        In-domain baseline & Robot only & Robot & 4.09 / 6.84 & 0.0099 / 0.0117 \\
        Mixed-domain & Robot + human data & Robot & 4.86 / 9.11 & 0.0096 / 0.0114 \\
        \bottomrule
    \end{tabular*}
\end{table}

\subsection{FK-Aware VLA Training}

The action-interface test verifies representability, but policy learning also requires the predicted joint actions to preserve task-space geometry. We therefore evaluate the network-level effect of FK-aware supervision in the PhysDex action head. All variants predict executable joint-space action chunks. The baseline is trained with joint-space action losses only, while the FK-aware variants add DS-HKC constraints that map the predicted joint chunk through the differentiable \robot kinematic model and supervise wrist and fingertip geometry in task space.

\begin{figure*}[t]
    \centering
    \includegraphics[width=0.95\textwidth]{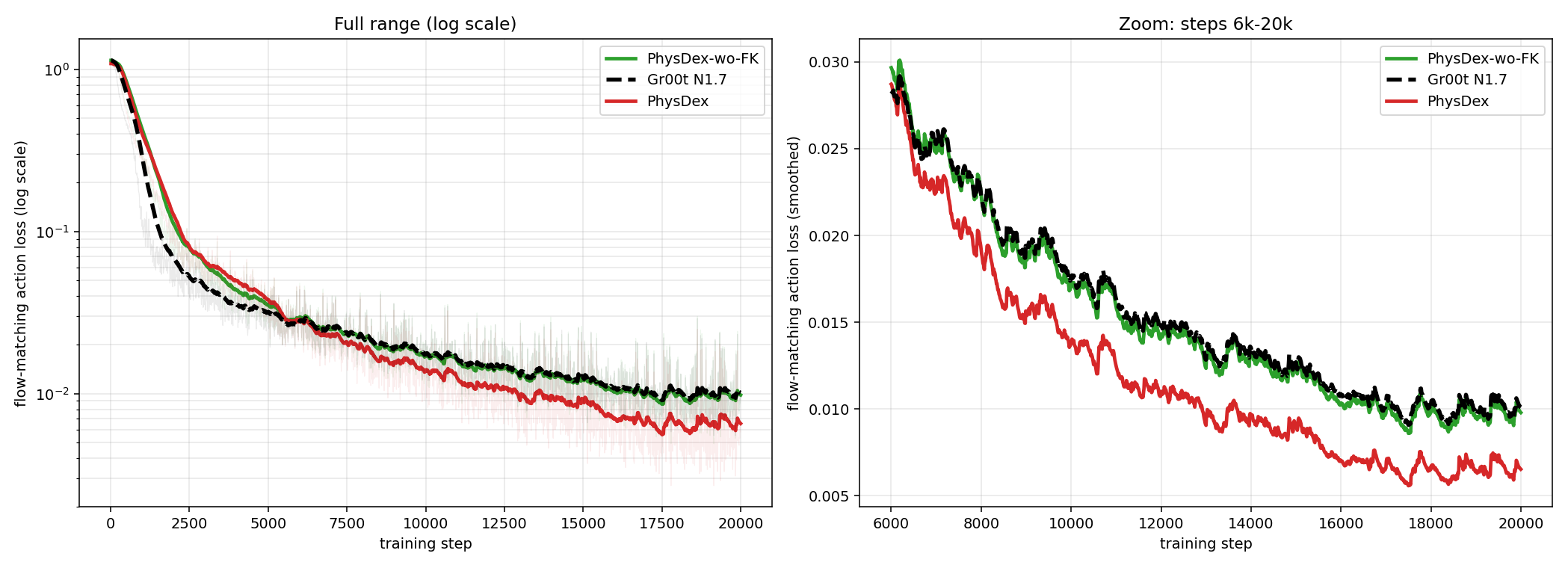}
    \caption{Training-loss comparison for high-DoF VLA action learning. FK-aware supervision reduces the loss by constraining the joint prediction through robot kinematics, and the PhysBrain-initialized FK-aware model reaches the lowest loss under the same training budget.}
    \label{fig:loss_storyline}
\end{figure*}

Figure~\ref{fig:loss_storyline} shows that adding FK supervision does not merely add an auxiliary penalty; it improves the optimization geometry of high-DoF action learning. A joint-only flow objective treats the 60 action dimensions largely as independent regression targets. In contrast, FK supervision couples the joints through the robot kinematic chain and measures whether the induced wrist and fingertip poses remain consistent with task-space manipulation geometry. The joint-limit term further discourages physically invalid solutions, and the warmup schedule lets the flow objective first learn coarse action alignment before the FK terms refine terminal geometry. This structured constraint reduces unconstrained trajectory memorization and yields smoother, lower loss curves, supporting the use of FK-aware training before real-robot deployment.

\subsection{Data Efficiency and Target-Task Deployment}

The final evaluation connects the validated Human-as-Humanoid conversion chain to robot-data efficiency and closed-loop PhysDex deployment. We compare against GR00T N1.7~\cite{GR00T_2025_arXiv} under the stage-wise protocol in Table~\ref{tab:task_stages}. For each task, we run 10 trials and report both ordered stage-completion rates and final task success. This protocol separates early perceptual or reaching failures from later dexterous failures, and therefore gives a more diagnostic comparison than a single binary success number.

\paragraph{Pretraining corpus and adaptation protocol.}
PhysDex uses a two-part initialization. The visual-language model is PhysBrain, a VLM trained on large-scale egocentric human data, which provides manipulation-aware visual-language features for objects, hands, contact cues, and task progress. The action autoencoding component is pretrained with our human-derived \dof labels. The pretraining corpus contains 1,500 hours of self-collected ego-exo human demonstrations covering diverse daily-life manipulation scenarios, and every sequence is converted into controller-aligned \dof robot action labels by the Human-as-Humanoid staged-IK pipeline. Thus the pretraining supervision is not only visual imitation data; it is high-DoF robot-action supervision in the target joint order, URDF convention, joint-limit range, and controller interface.

Downstream adaptation is evaluated in two regimes. For ring placement, magic-cube packing, cup stacking and water pouring, post-training uses target-task human demonstrations only; these tasks therefore test zero target-task robot-demonstration generalization. For temperature-gun measurement, light-bulb loosening, and bottle-cap loosening, sustained contact and fine hand-object interaction make the tasks more sensitive to hardware grounding. We therefore use a small amount of real-robot data for mid-training anchoring and task-specific post-training, following the EgoScale training recipe~\cite{Egoscale_2026_arXiv}. GR00T N1.7 is adapted under the same downstream data setting: no target-task robot data for the human-only tasks, and the same real-robot data budget for the robot-assisted tasks.

\begin{figure*}[p]
    \centering
    \includegraphics[width=0.98\textwidth,height=0.88\textheight,keepaspectratio]{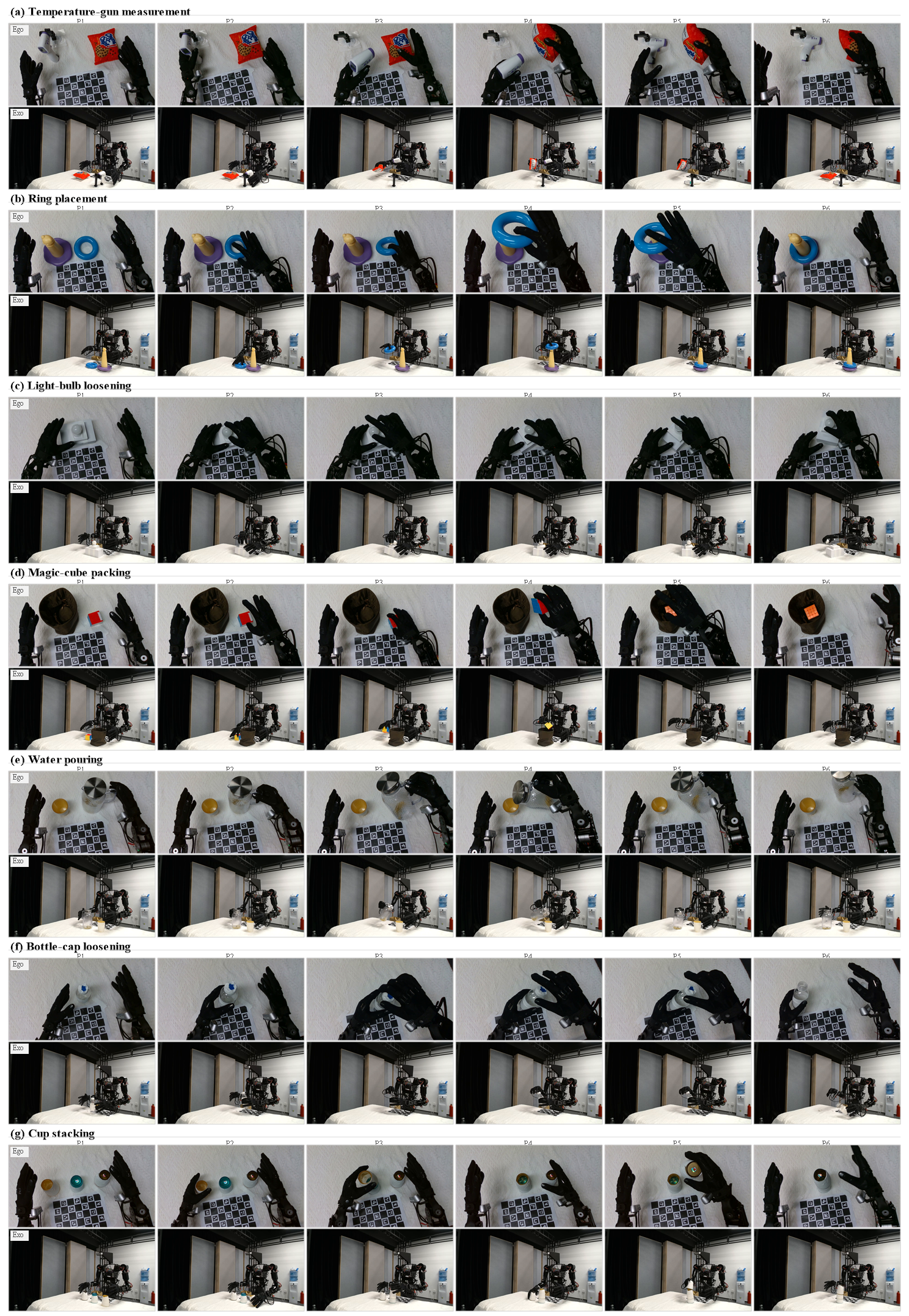}
    \caption{Synchronized rollout frames for downstream manipulation tasks. Each panel corresponds to one task; within each panel, the upper row shows the egocentric policy observation and the lower row shows a third-person visualization view at matched phases from early to late execution.}
    \label{fig:ego_exo_task_rollouts}
\end{figure*}

Figure~\ref{fig:ego_exo_task_rollouts} provides synchronized rollout frames for the downstream task suite, showing the policy-observation view together with a third-person view used only for qualitative visualization of real-robot execution.

\begin{figure*}[t]
    \centering
    \includegraphics[width=0.95\textwidth]{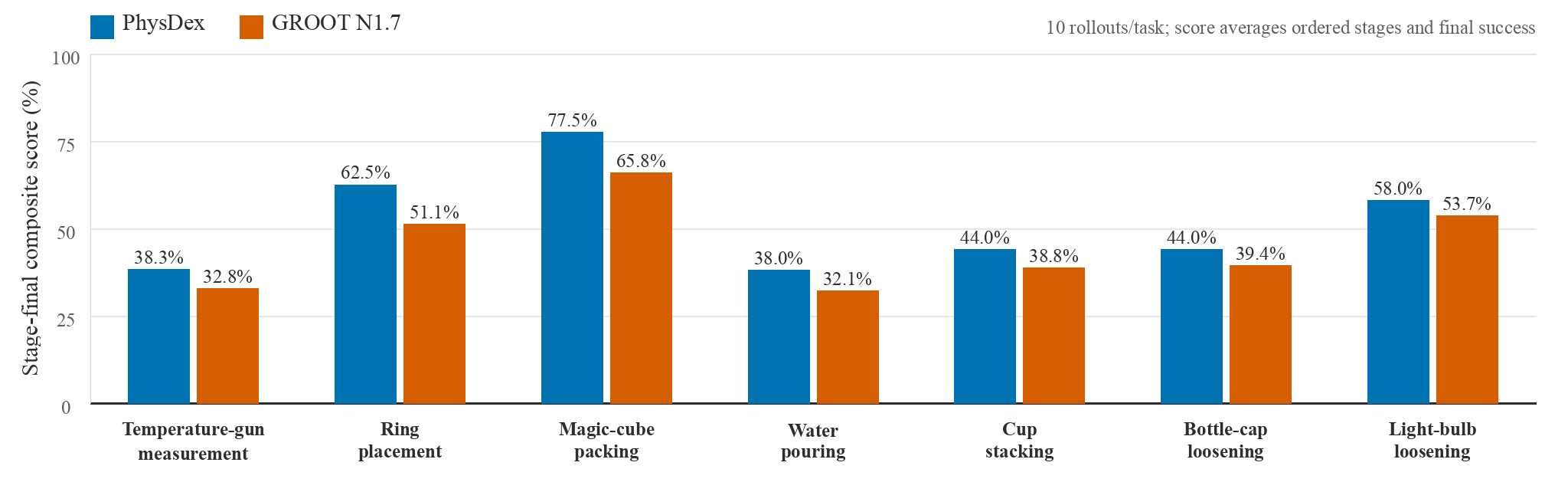}
    \caption{Stage-final composite score in real-robot deployment. The score aggregates ordered stage completion and final task success over 10 rollouts per task, providing a progress-aware measure of deployment performance rather than a binary final-success rate alone.}
    \label{fig:vla_final_success}
\end{figure*}

This protocol tests two questions at deployment time. The human-only tasks evaluate whether Human-as-Humanoid labels support PhysDex target-task transfer without robot demonstrations. The robot-assisted tasks evaluate whether the 1,500-hour human-derived pretraining reduces the amount of real robot data needed when precise contact and hardware grounding are required. Across both regimes, the comparison between PhysDex and GR00T N1.7 measures stage-wise task progress and final deployment success under the same rollout budget and matched downstream data condition.

Figure~\ref{fig:vla_final_success} summarizes the preliminary stage-final composite results over seven real-robot tasks. The score combines ordered stage completion with final task success, providing a progress-aware measure of deployment performance rather than a binary success indicator alone. PhysDex obtains higher composite scores than GR00T N1.7 across the evaluated tasks. The improvement is more pronounced in the human-only adaptation regime, where target-task training uses converted human demonstrations without robot demonstrations, and is smaller in the robot-assisted regime, where both methods receive the same limited amount of real-robot data. This trend is consistent with the role of Human-as-Humanoid labels: the converted data provides embodiment-specific, controller-aligned high-DoF action supervision that directly matches the \robot action interface. GR00T N1.7 remains a strong generalist humanoid baseline, but this evaluation stresses target-robot-specific dexterous action learning under limited robot data, where additional human-derived action supervision provides a more specialized action prior. Since each task currently uses 10 rollouts, we interpret the result as a controlled preliminary comparison rather than a final statistical claim.

Together, the deployment comparison, action-tokenizer reconstruction, and FK-aware loss analysis support the same conclusion: human demonstrations are not only useful as visual pretraining data, but can also serve as a stable source of executable action supervision for VLA scaling. Robot data remains important for defining the controller interface, grounding the policy to hardware, and refining contact-rich edge cases, but scalable human-derived supervision reduces the amount of target-task robot data that the policy must rely on.


\section{Discussion and Limitations}
\label{sec:discussion}

Our results support near-real-time human-to-humanoid action generation as a practical route for scaling humanoid action data, but several limitations remain. First, pose-estimation quality bounds downstream retargeting quality: systematic pose-estimation failures can become action-label bias. Second, IK quality bounds policy quality because the human-derived action data inherits the robot model, joint limits, and calibration used by the retargeter. Third, the current pipeline is tied to a specific robot URDF and joint convention; transferring to a new embodiment requires retargeting and adapting the action dimension. Fourth, human-derived actions capture kinematics more directly than contact forces, so robot data remains important for anchoring, evaluation, and final adaptation in contact-rich settings. This limitation is especially pronounced for fine-grained dexterous manipulation. Human-derived labels primarily provide kinematic supervision, while tasks such as cap loosening, bulb loosening, and button pressing depend on contact force, friction, local slip, fingertip placement, and hand-object morphology. Small errors in retargeted finger pose or contact geometry can change the task outcome, and the morphology gap between the human hand and the robot dexterous hand cannot be eliminated by skeleton alignment alone. Zero target-task robot demonstrations are therefore more feasible for geometry-dominant or moderate-contact tasks, while high-precision dexterous tasks still benefit from limited real-robot data for hardware grounding. Our zero-shot claim therefore refers to target-task deployment without target-task robot demonstrations, not to eliminating all robot-specific modeling assumptions.

These limitations clarify rather than weaken the central claim that human data can improve the data efficiency of VLA scaling. The point is not to replace robot data, but to avoid asking target-task robot demonstrations to carry all of the scale. Human data can provide broad, diverse, executable action supervision through collection-time keypoint estimation and IK retargeting; robot-specific resources can then be concentrated where they have the highest value: defining the kinematic interface, grounding the policy to hardware, measuring deployment performance, and calibrating the remaining embodiment gap.

In future work, we will further relax the data-capture assumptions of Human-as-Humanoid. The current pipeline uses synchronized ego-exo videos because exocentric views provide stable evidence for upper-body and hand recovery. We will study accurate egocentric skeleton and hand-keypoint recovery under occlusion, motion blur, and camera motion. Beyond recovering motion from recorded videos, we will also study egocentric hand-object interaction generation: by explicitly specifying interaction objects, task goals, and hand/object trajectories, the generator could synthesize temporally stable and physically plausible first-person interaction videos. Coupled with the Human-as-Humanoid conversion chain, such generated interaction data could be transformed into a larger corpus of humanoid-executable action supervision for high-DoF VLA pretraining.


\section{Conclusion}
\label{sec:conclusion}

High-DoF humanoid VLA learning requires scalable observation--action supervision that is both diverse and executable on the target robot.
We presented \method, a human-to-humanoid data-processing pipeline that converts synchronized ego-exo human demonstrations into controller-aligned \dof action chunks on the human-aligned \robot embodiment.
These labels make human videos usable for PhysDex, which predicts executable joint-space actions while preserving wrist and fingertip geometry through FK-aware constraints.
Experiments validate the conversion chain across motion recovery, robot-action-space compatibility, and real-robot deployment, showing a 4.8--7.2x raw throughput gain over humanoid teleoperation and successful deployment without target-task robot demonstrations on several tasks.
Overall, the results suggest that human-derived executable supervision can reduce the robot-data burden for humanoid VLA scaling, while robot data remains valuable for hardware grounding and contact-rich refinement.


\section*{Acknowledgments}
We gratefully acknowledge the DeepCybo hardware team for their substantial support in the development and integration of the humanoid platform, with particular thanks to Xuguo HE, Qiyuan SU, Hong LI and Haochen LIU. We also acknowledge Wuji for their technical support on the dexterous-hand system.

\bibliographystyle{assets/plainnat}
\bibliography{example}

\end{document}